\documentclass[sigconf]{acmart}
\usepackage{caption, subcaption}
\usepackage{fancyvrb}
\usepackage{multirow} 
\usepackage{lineno}
\usepackage{varwidth}

\AtBeginDocument{%
  \providecommand\BibTeX{{%
    \normalfont B\kern-0.5em{\scshape i\kern-0.25em b}\kern-0.8em\TeX}}}

\setcopyright{acmcopyright}
\copyrightyear{2018}
\acmYear{2018}
\acmDOI{XXXXXXX.XXXXXXX}

\acmConference[Conference acronym 'XX]{Make sure to enter the correct   conference title from your rights confirmation emai}{June 03--05,   2018}{Woodstock, NY}
\acmPrice{15.00}
\acmISBN{978-1-4503-XXXX-X/18/06}

\setcopyright{none}
\renewcommand\footnotetextcopyrightpermission[1]{} % removes footnote with conference information in first column
\begin{document}
\title{Deep dive into language traits of AI-generated Abstracts \vspace{1cm}}

\author{Vikas Kumar}
\email{vikas@cs.du.ac.in}
\orcid{0000-0001-9882-7310}
\affiliation{%
  \institution{University of Delhi}
  %\streetaddress{P.O. Box 1212}
  \city{New Delhi}
  \country{India}
  \vspace{1cm}
}

\author{Amisha Bharti}
\email{abharti@cs.du.ac.in}
\orcid{0009-0008-4018-6117}
\affiliation{%
  \institution{University of Delhi}
  %\streetaddress{P.O. Box 1212}
  \city{New Delhi}
  \country{India}
}

\author{Devanshu Verma}
\email{dverma@cs.du.ac.in}
\orcid{0009-0009-3526-2338}
\affiliation{%
  \institution{University of Delhi}
  %\streetaddress{P.O. Box 1212}
  \city{New Delhi}
  \country{India}
}

\author{Vasudha Bhatnagar}
\email{vbhatnagar@cs.du.ac.in}
\orcid{0000-0002-9706-9340}
\affiliation{%
  \institution{University of Delhi}
  %\streetaddress{P.O. Box 1212}
  \city{New Delhi}
  \country{India}
}

\renewcommand{\shortauthors}{Kumar et al.}

\newcommand{\vikasR}[1]{\textcolor{red}{#1}}
\newcommand{\vikasG}[1]{{#1}}

\begin{abstract}
%Human like writing by chat GPT has led to generation of scientific scholarly texts, showcased in recent conferences and journals.  Automatic detection of these text is challenging due to the length of these texts.  In this work we attempt to detect the \textit{Abstracts}  generated by chat GPT, which are much shorter in  length and \textbf{bounded}.  We extract a set of semantic and  lexical properties of the text, and observe that non-linear classifiers can confidently detect these Abstract.

Generative language models, such as ChatGPT,  have garnered attention for their ability to generate human-like writing in various fields, including academic research. The rapid proliferation of generated texts has bolstered the need for automatic identification to uphold transparency and trust in the information. However, these generated texts closely resemble human writing and often have subtle differences in the grammatical structure, tones, and patterns, which makes systematic scrutinization challenging. In this work, we attempt to detect the \textit{Abstracts}  generated by ChatGPT, which are much shorter in length and \textbf{bounded}.  We extract the text's semantic and lexical properties and observe that traditional machine learning models can confidently detect these Abstracts.
\end{abstract}





\begin{teaserfigure}
  % \includegraphics[width=\textwidth]{}
  % \caption{Seattle Mariners at Spring Training, 2010.}
  \Description{Enjoying the baseball game from the third-base
  seats. Ichiro Suzuki preparing to bat.}
  \label{fig:teaser}
\end{teaserfigure}


\maketitle

\section{Introduction}
The Abstract of a scientific scholarly article is a \textit{meta-discourse} consisting of explicit linguistic markers and expressions that enable authors to summarize their own discourse. It plays a crucial role in conveying the intent of the research and empowers the reader to decide whether the article is of interest or not. A well-written abstract is a coherent and cohesive text that conveys the objective(s) of the research, a brief background, and research conclusions in a limited number of words. Due to the confident and fluent text generated by ChatGPT\footnote{\url{https://openai.com/}} (CG), researchers have recently used it for generating scientific scholarly text. Several journal and conference publications authored by CG have been showcased in recent times \cite{king2023conversation,askr2023future,transformer2022rapamycin}. However, scientists have raised serious concerns about the use of ChatGPT in science \cite{sallam2023chatgpt,thomas2023grappling} %\cite{else23abstract-fools} 
and education \cite{arif2023future, lo2023impact}.  

%else23abstract-fools --> ABSTRACTS WRITTEN BY CHATGPT FOOL SCIENTISTS

Fluency in the language generated by the large language models is astounding and it challenges discrimination between human-written and AI-generated texts. Early research endeavors, driven by curiosity, have revealed differences in linguistic patterns in human-written and CG text in Q-A \cite{guo2023close}, generic \cite{dou2021gpt,johansson2023tale,mitchell2023detectgpt} texts. Due to the demonstrated use of AI in successfully generating scientific scholarly text \cite{king2023conversation, askr2023future}, it is now a necessity to develop tools to identify AI-generated content in submitted papers and detect unethical practices in paper writing. Research in this direction serves the scientific society and aims to  minimize reliance on Large Language Models (LLMs) for scientific thinking \cite{else2023abstracts}. On the other side of the same coin, it benefits the Natural Language Processing (NLP) community by pointing out the gaps that need to be filled in LLMs to improve the quality of AI-generated text.

%2023-is-gpt3-text-indistinguishable -->  Is GPT-3 Text Indistinguishable from Human Text? SCARECROW: A Framework for Scrutinizing Machine Text
% some-papaers-with-cg-author ---> ADD 3 REFERENCES ---> RFC 9405 AI Sarcasm Detection: Insult Your AI without Ofending It, Can a Transformer Assist in Scientific Writing? Generating Semantic Web Paper Snippets with GPT-2, A Primer on Deep Reinforcement Learning for Finance
% %  2023-tale-of-two-texts    ---> A Tale of Two Texts, a Robot, and Authorship: a comparison between a human-written and a ChatGPT-generated text
% 2023-jan-zeroshot ---> DetectGPT: Zero-Shot Machine-Generated Text Detection using Probability Curvature

%The number of papers on Google scholar with  \textit{ChatGPT} in title is a testimony to the importance attached by the NLP researchers to this task. IN this context, we present a quick analysis of the content indexed by GS\footnote{ Content as on 6 July 2023} in Table \ref{cg-papers-stats}. 

Recent interest in the detection of AI-generated abstracts of scientific papers is reflected in  \cite{ma2023abstract, gao2023comparing}. Gao et al. gathered fifty research abstracts from high-impact factor medical
journals and asked ChatGPT to generate research abstracts based on their titles and journals \cite{gao2023comparing}. Subsequently, they found that while blinded human reviewers could correctly identify 68\% of generated abstracts and  86\% of original articles, "GPT-2 Output Detector"\footnote{\url{https://huggingface.co/openai-detector.}} could detect  AI-generated output with desirably high accuracy (AUROC = 0.94). Ma et al. investigate the gap between Abstracts generated by GPT3 and those written by humans by using writing-style features \cite{ma2023abstract}.  Based on the GPT-2 output detector, fine-tuned on the collected dataset, they conclude that AI-generated text is more likely to contain errors in language redundancy and facts. 

We present a systematic study of the linguistic features of the \textit{Abstracts} of scientific scholarly texts generated by {\itshape ChatGPT} using classical machine learning models, with two-fold objective. First, we aim to understand the features that are important for discriminating between human-written and generated \textit{Abstracts}. The choice of feature-based discrimination is motivated by the findings of Ma et al. \cite{ma2023abstract}, who find merit in feature-based discrimination. We use three pragmatic devices (Hedgers, Boosters, and Hype-words) that are commonly used in scientific scholarly writings \cite{poole2019epistemic}, in addition to semantic and surface-level features of the text. Second, in the current era of expensive and energy-guzzling deep neural models in NLP research, we explore the efficacy of (forgotten) established models in machine learning repertoire for text classification. These two objectives construe the contribution of our work.

We describe \textit{CHEAT} dataset \cite{yu2023cheat}, and two small curated datasets of scientific abstract in Sec. \ref{sec:datasets}, discriminating semantic and pragmatic properties of the text in Sec. \ref{sec:features}, and five traditional classifiers in Sec. \ref{sec:classifiers}, We report experimental results in Sec. \ref{sec:results}. 
\section{Methodology}
We describe in detail the steps followed for this empirical study. The following subsections elucidate the data collection, feature extraction and evaluation protocol. 
\subsection{Data Set}
\label{sec:datasets}
We curate the dataset in two phases. First, we select \textbf{thirty research articles }(2014 - 2021)\footnote{We restricted the year of publication to 2021 to align with the limit of the training data of ChatGPT.}  (co-)authored by three researchers personally known to the authors. These articles may probably were seen by the model during training. We extract the author abstracts (AA) and the author keywords from the papers and   generate five ChatGPT Abstract (CGA) abstracts for each paper using the prompt shown below. 
\vspace{0.2cm}
\noindent
 \fbox{ \begin{minipage}{0.95\linewidth}
\textit{ChatGPT Prompt: Write a scientific abstract for the paper entitled} \texttt{PAPER\_TITLE} \textit{using the keywords} \texttt{KEYWORDS}. \textit{Response length must contain at least} \texttt{NO\_OF\_WORDS} \textit{tokens}.
\end{minipage}   
 } 
%\vspace{0.2cm}
 The prompt is designed to be simple, yet precise and concise to convey the research intent of the author(s).  Multiple interactions with ChatGPT using the same  prompt  reveal the consistency and semantic coherence in the generated \textit{Abstracts}, which can be evaluated  by the authors.  Each \textit{Abstract} for a paper is generated in a fresh session after clearing the context for enhanced user control and  to enable assessment solely on the basis of the current prompt, without any influence from the previous one. We call this as \textit{Human-Eval }dataset.

In the second phase, we retrieve papers from Google Scholar using "ChatGPT" in query, with more than 25 citations.  After filtering, we retained 82 papers, from which we extract Author abstracts (AA) and titles. We generate one  ChatGPT abstract (CGA) using the same prompt, sans the author-supplied keywords. With this prompt, ChatGPT gets more freedom to construe the context and  generate the text according to the title in contrast to \textit{Human-Eval} dataset, where the generative space of the model is restrictive due to the given keywords.  Please note that these abstracts have definitely not been seen by the model during training. We refer to this data set as \textit{CG-Articles} dataset.  

Comparative statistics of three data sets are shown in Table \ref{tab:dataset-comparison}. Numbers below the datasets indicate the total number of \textit{Abstracts}. Numbers in the table are the ratios of the corresponding metrics observed in the author-written abstracts to the synthetic abstracts. We observe a similar trend in ratios for all three datasets.
%

%%%%%%%%%%%%%%%%%%%%%%%% FIGURE SHOWING DISTRIBUTION %%%%%%%%%%%%%%%%%%%%%%%%%%%
\begin{figure*}[ht]
  \centering
  %\adjustbox{max width=\textwidth}{
      \begin{subfigure}{0.33\textwidth}
        \includegraphics[width=\textwidth, height = 1.7in]{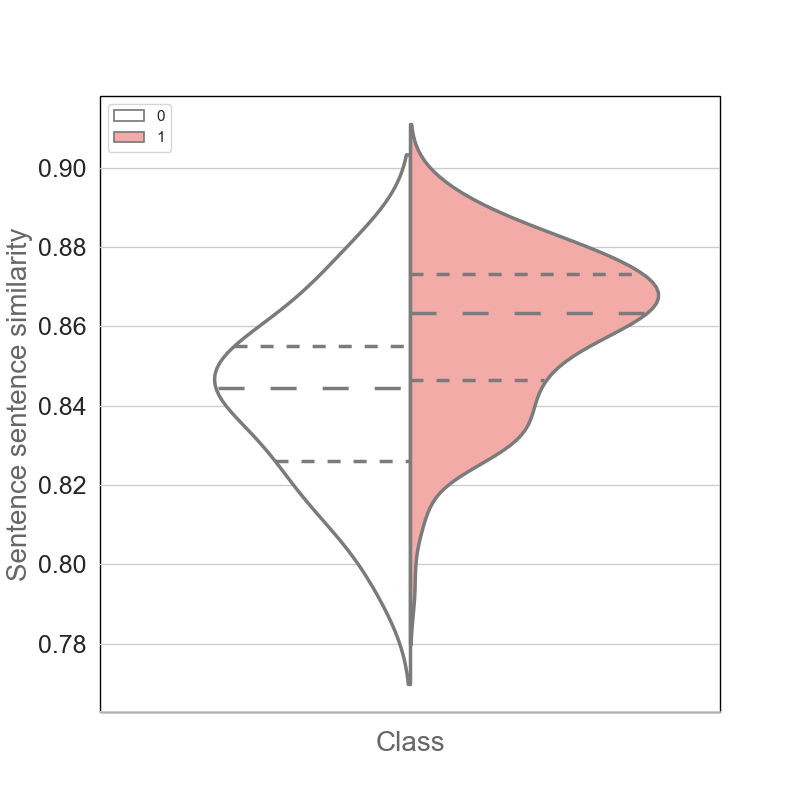}
        \caption{Sentence-sentence similarity}
        \label{fig:As_1p}
    \end{subfigure}
    \begin{subfigure}{0.33\textwidth}
        \includegraphics[width=\textwidth, height = 1.7in]{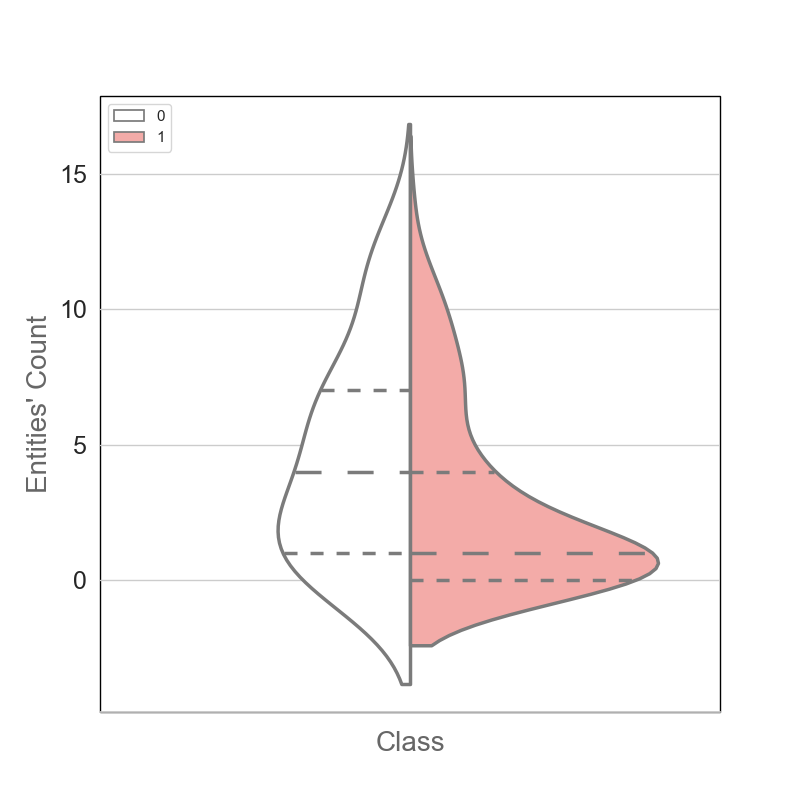}
        \caption{Entities' Count}
        \label{fig:As_2p}
    \end{subfigure}
    \begin{subfigure}{0.33\textwidth}
        \includegraphics[width=\textwidth, height = 1.7in]{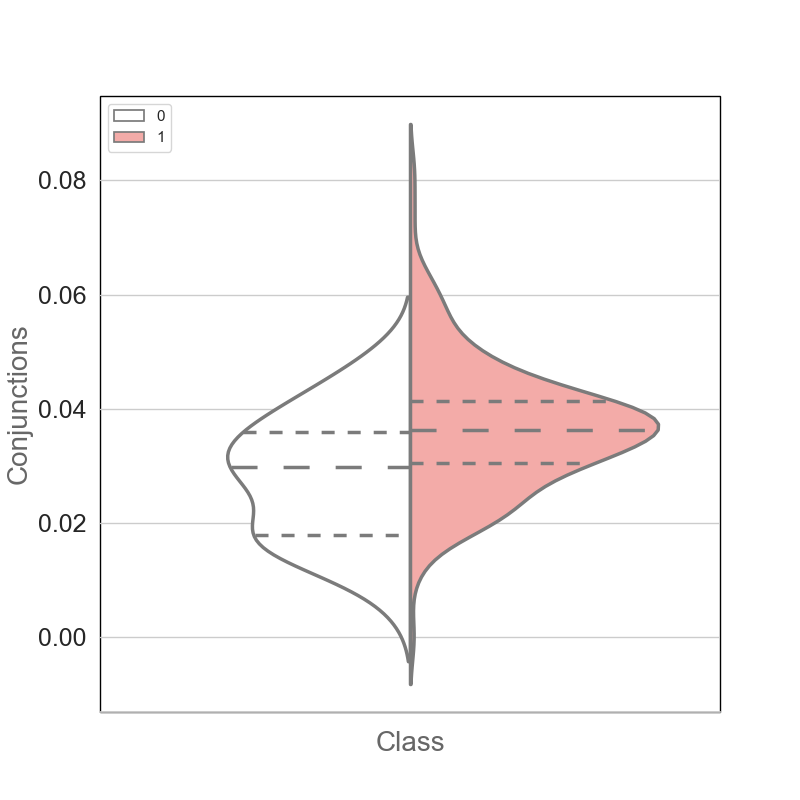}
        \caption{Conjunctions}
        \label{fig:As_3p}
    \end{subfigure} 
    %}
  \caption{Categorical distribution of handpicked features in Human-Eval dataset.}
  \label{fig:fig-1}
\end{figure*}

% _______________________
\begin{figure*}[ht]
  \centering
  %\adjustbox{max width=\textwidth}{
      \begin{subfigure}{0.33\textwidth}
        \includegraphics[width=\textwidth, height = 1.5in]{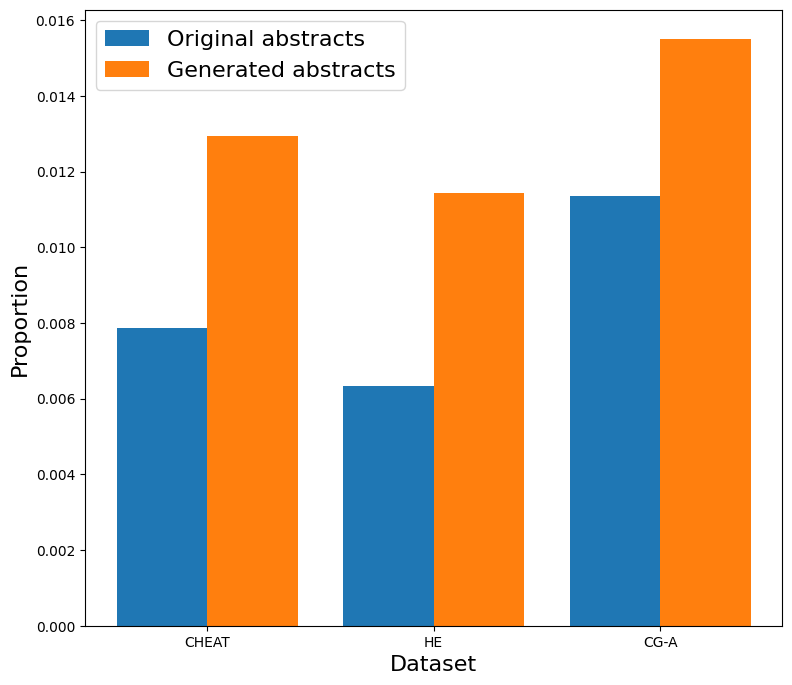}
        \caption{Hyping lemma}
        \label{fig:hb_1p}
    \end{subfigure}
    \begin{subfigure}{0.33\textwidth}
        \includegraphics[width=\textwidth, height = 1.5in]{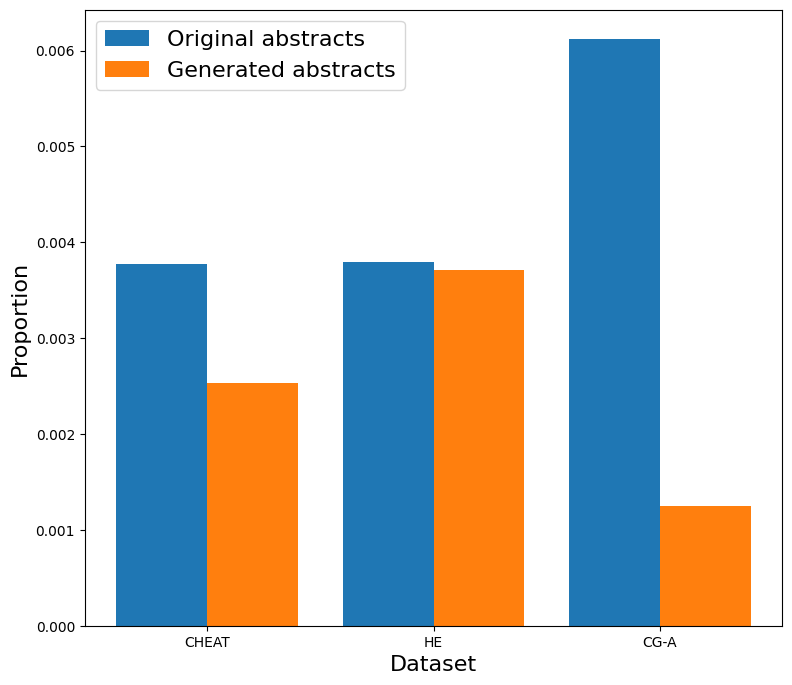}
        \caption{Hedge words}
        \label{fig:hb_2p}
    \end{subfigure}
    \begin{subfigure}{0.33\textwidth}
        \includegraphics[width=\textwidth, height = 1.5in]{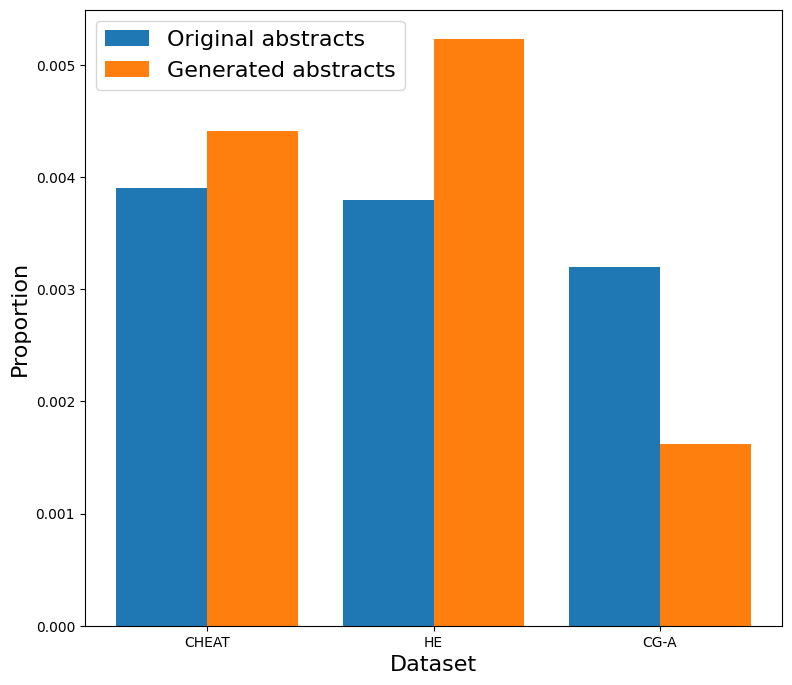}
        \caption{Booster words}
        \label{fig:hbs_3p}
    \end{subfigure} 
    %}
  \caption{Class-wise distribution comparison of hyping lemma, hedge words and booster words between ChatGPT and human written abstracts for considered dataset.}
  \label{fig:fig-2}
\end{figure*}
% _______________________

\begin{table}[h]
 \caption{Ratio of the metrics for three datasets. HE: \textit{Human-Eval} dataset, CG-A: \textit{CG-Articles} dataset, CHEAT \cite{yu2023cheat} dataset. Each number is the ratio of the corresponding metric in Author abstract to ChatGPT Abstract. }
\begin{tabular}{|l|l|l|l|}
\hline
  Metric        & HE & CG-A & CHEAT \\
   & (198)   & (82)      & (30790)       \\ \hline \hline
\# Sentences    &1.0560 &	1.0657	& 1.2331        \\ \hline
\# Words        & 1.0219 &	1.0847&	1.2286      \\ \hline
\# Unique Words & 1.1520 &	0.9623&	1.1007    \\ \hline
\# Stopwords    & 1.0453 &	1.2517&	1.0039     \\ \hline
Sentence Length & 0.97    & 1.0242     & 1.0034 \\ \hline

\end{tabular}
 \label{tab:dataset-comparison}
\end{table} 

 \textit{CHEAT} dataset, proposed in \cite{yu2023cheat} is used for training the predictive model, while these two small data sets are used as test sets. Since chat GPT is available only through the website, we manually generated the abstracts, which explains the small size of the two curated datasets. We plan to scale-up the two datasets in near future and make them public. The contrast in the timeline of \textit{Human-Eval} and \textit{CG-Articles} datasets is expected to make them valuable resources for research and development of LLMs.
\subsection{Feature Extraction and Analysis}
\label{sec:features}
%Extraction and Selection
The problem addressed in this work is similar to the author-attribution problem. Acknowledging that ChatGPT generates fluent and confident language, we refrain from scrutinizing grammatical features in two categories of abstract. Instead, we focus on linguistic, semantic and pragmatic characteristics of the texts and scrutinize their potency to discriminate.  

%This launch an inquiry into pragmatics of the discourse to obtain insights about the text properties that are possibly capable of discriminating between AA and CGA.

\subsubsection{Linguistic Features (LF)}
  Following Yu et al. \cite{yu2023cheat},  we scrutinize the family of semantic, coherence and cohesion features of the short discourse generated by {\itshape ChatGPT} and compare with those extracted from author-written abstracts.  TAACO Tool\footnote{\url{https://www.linguisticanalysistools.org/taaco.html}} is the outcome of recent advances in computational discourse analysis by quantifying the quality of text using a wide spectrum of grammatical devices.
  We use LSA feature of the tool to extract document and sentence level features, and sentence-sentence interaction features \cite{crossley2019tool, crossley2016tool}. We consciously ignored paragraph-paragraph level features since most abstracts contain only one paragraph.  Some of these features measure local (sentence level) and global (document level) cohesion. In all, we extract 102 linguistic  features\footnote{\label{git_link}https://github.com/DUCS-23/CGA} from the text.

%%%%%%%%%%%%%%%%%%%%%%%%%%%%%%%%%%%%%%%%%%%%%%%%%%%%%%%%%%%%%%%%%%%%%%%%%%%%%%%%%%%%%%%

\subsubsection{Semantic Features (SF)}
We examine the semantics from multiple viewpoints, including text cohesion, semantic similarity and semantic entities. Cohesion is a semantic property of the text that connotes smooth integration of the entities in the text. It is an objective property that is quantified by extracting the semantic relationship among words or phrases in the text. Cohesive cues present in the text aid the reader in process of comprehension so that the entities and concepts in the text can be connected correctly. They are indicated by linguistic features such as lexical repetition and anaphora, which are explicitly realized at the surface level of the text. The feature is discriminating because the author desires to compress as many ideas in the abstract as possible and hence prioritizes them, thereby minimizing repetition and lowering cohesion. ChatGPT on the other hand, faces no such compulsion and generates text algorithmically in a probabilistic manner, prioritizing the smoothness of text over the research exposition. %The choice of cohesion features is motivated by the conjecture that the abstract generated by the trained language model will exhibit different degree of cohesion than those written by  the scientist.

Semantic features include sentence-sentence similarity using SciBert \cite{Beltagy2019SciBERT}, sentence-title similarity and named entities in the text.  In all, we compute ten\footref{git_link} semantic features from the texts. Fig. \ref{tab:dataset-comparison} contrasts the distribution of three handpicked semantic features for two classes in\textit{ Human-Eval } dataset. 
\subsubsection{Pragmatic Features}
We use three prominent pragmatic devices that modify the author's intent by introducing uncertainty, strengthening or emphasizing a claim,  thereby influencing the interpretation of the intended meaning by the reader. Use of \textit{Hedge} words, \textit{Boosters} and \textit{Hyping-lemmas} (HBH) are devices used in scientific texts to introduce positivity bias and convey authors' stance and attitude towards the information they are presenting. A recent study by Yao et al. based on articles in the last 25 years reveals a clear shift in the use of these devices in scientific writing \cite{yao2023promoting}. Fig \ref{fig:fig-2} shows the distribution of these devices in texts for all three datasets. Again, we find largely similar trends across the datasets.

%%%%%%%%%%%% CHEAT FOR TRAINING, HUMAN-EVAL FOR TESTING %%%%%%%%%%%%
%
\begin{table*}[ht!]
    \caption{ 5-fold Cross-validation performance (F1 Score) of Machine Learning Models trained on CHEAT dataset. }
    %\adjustbox{max width=\textwidth}{
    \begin{tabular}{|l|llll|llll|}
    \hline
    Classifier & \multicolumn{1}{c|}{\begin{tabular}[c]{@{}c@{}} SF \\ (10)\end{tabular}} & \multicolumn{1}{c|}{\begin{tabular}[c]{@{}c@{}}LF\\ (102)\end{tabular}} & \multicolumn{1}{c|}{\begin{tabular}[c]{@{}c@{}}SF + LF + HBH \\ (115)\end{tabular}} & \multicolumn{1}{c|}{\begin{tabular}[c]{@{}c@{}}Sel. Features\\ (5)\end{tabular}}  & \multicolumn{1}{c|}{\begin{tabular}[c]{@{}c@{}}Sel. Features\\ (10)\end{tabular}} & \multicolumn{1}{c|}{\begin{tabular}[c]{@{}c@{}}Sel. Features\\ (15)\end{tabular}} & \multicolumn{1}{c|}{\begin{tabular}[c]{@{}c@{}}Sel. Features\\ (20)\end{tabular}} & \multicolumn{1}{c|}{\begin{tabular}[c]{@{}c@{}}Sel. Features\\ (25)\end{tabular}} \\ \hline
            & \multicolumn{1}{c|}{\scriptsize{Col. 1}}   & \multicolumn{1}{c|}{\scriptsize{Col. 2}}  & \multicolumn{1}{c|}{\scriptsize{Col. 3}} & \multicolumn{1}{c|}{\scriptsize{Col. 4}} & \multicolumn{1}{c|}{\scriptsize{Col. 5}} & \multicolumn{1}{c|}{\scriptsize{Col. 6}} & \multicolumn{1}{c|}{\scriptsize{Col. 7}} & \multicolumn{1}{c|}{\scriptsize{Col. 9}}  \\ \hline
    LDA        & \multicolumn{1}{c|}{0.8390 }   & \multicolumn{1}{c|}{0.8797 }  & \multicolumn{1}{c|}{0.9129 } & \multicolumn{1}{c|}{0.8227 } & \multicolumn{1}{c|}{0.8609 } & \multicolumn{1}{c|}{0.8777 } & \multicolumn{1}{c|}{ 0.8842} & \multicolumn{1}{c|}{0.8896 }  \\ \hline
    LR        & \multicolumn{1}{c|}{0.8476 }   & \multicolumn{1}{c|}{0.8811}  & \multicolumn{1}{c|}{0.9228 } & \multicolumn{1}{c|}{ 0.8289}   & \multicolumn{1}{c|}{ 0.8727}    & \multicolumn{1}{c|}{0.8861 }  & \multicolumn{1}{c|}{ 0.8919}   & \multicolumn{1}{c|}{0.8958}  \\ \hline
    SVC         & \multicolumn{1}{c|}{ \hspace{0.2cm} 0.8744\hspace{0.2cm}  }   & \multicolumn{1}{c|}{\hspace{0.2cm} 0.8956 \hspace{0.2cm}  }  & \multicolumn{1}{c|}{ 0.9328} & \multicolumn{1}{c|}{ 0.8618}   & \multicolumn{1}{c|}{0.8982 }    & \multicolumn{1}{c|}{ 0.9078}  & \multicolumn{1}{c|}{ 0.9132}   & \multicolumn{1}{c|}{0.9175}  \\ \hline
    XGB         & \multicolumn{1}{c|}{0.8813 }   & \multicolumn{1}{c|}{ 0.8930}  & \multicolumn{1}{c|}{0.9316} & \multicolumn{1}{c|}{0.8750 }   & \multicolumn{1}{c|}{0.9040 }    & \multicolumn{1}{c|}{ 0.9097}  & \multicolumn{1}{c|}{ 0.9168}   & \multicolumn{1}{c|}{0.9204}  \\ \hline
    ETC  & \multicolumn{1}{c|}{ 0.8730}   & \multicolumn{1}{c|}{0.8490 }  & \multicolumn{1}{c|}{0.9011 } & \multicolumn{1}{c|}{0.8612 }   & \multicolumn{1}{c|}{0.8978 }    & \multicolumn{1}{c|}{ 0.9018}  & \multicolumn{1}{c|}{0.9056 }   & \multicolumn{1}{c|}{0.9075}  \\ \hline
    \end{tabular}%}
    \label{tab:cheatAUC}
    \end{table*}

\subsection{Classifiers \& Evaluation}
\label{sec:classifiers}
  We select five prominent classification algorithms including both linear and non-linear models. Specifically, we use  Linear Discriminant Analysis (LDA)~\cite{fisher1936use}, Logistic Regression (LR)~\cite{cox1989analysis}, Support Vector Classifier (SVC)~\cite{vapnik1999nature}, XGBoost (XGB)~\cite{chen2016xgboost} and Extra Tree Classifier (ETC)~\cite{geurts2006extremely} as representative algorithms for our analysis.  The \textit{scikit-learn} and \textit{xgboost} python packages implementation is used with default hyperparameters for each representative algorithm. We employ $F_1$ score and Area Under Curve (AUC) as evaluation metrics to assess the predictive performance. 
\section{Results and Discussion}
\label{sec:results}
We present the results of the empirical study. We first analyse the feature set and study the performance of the five classification algorithms, and then discuss feature selection. We report an ablation study on the three types of features (linguistic, semantic, and pragmatic) and find that a small subset of features is able to achieve performance comparable to the one reported in the \cite{yu2023cheat}.

    \begin{table*}[ht!]
    \caption{Test performance of Machine Learning Models trained on CHEAT dataset. }
    %\adjustbox{max width=\textwidth}{
    \begin{tabular}{|l|llll|llll|}
    \hline
    Dataset & \multicolumn{4}{c|}{HUMAN-EVAL} & \multicolumn{4}{c|}{CG-Articles }\\ \hline
    \multirow{2}{*}{Classifier} & \multicolumn{2}{c|}{\begin{tabular}[c]{@{}c@{}}SF + LF + HBH (115)\end{tabular}} & \multicolumn{2}{c|}{\begin{tabular}[c]{@{}c@{}}Sel. Feature (25)\end{tabular}} & \multicolumn{2}{c|}{\begin{tabular}[c]{@{}c@{}}SF + LF + HBH (115)\end{tabular}} & \multicolumn{2}{c|}{\begin{tabular}[c]{@{}c@{}}Sel. Features (25)\end{tabular}}  \\ \cline{2-9}
    &\multicolumn{1}{c|}{F1}&\multicolumn{1}{c|}{AUC}&\multicolumn{1}{c|}{F1}&\multicolumn{1}{c|}{AUC}&\multicolumn{1}{c|}{F1}&\multicolumn{1}{c|}{AUC}&\multicolumn{1}{c|}{F1}&\multicolumn{1}{c|}{AUC}\\ 
    \hline

                & \multicolumn{1}{c|}{\scriptsize{Col. 1}}   & \multicolumn{1}{c|}{\scriptsize{Col. 2}}  & \multicolumn{1}{c|}{\scriptsize{Col. 3}} & \multicolumn{1}{c|}{\scriptsize{Col. 4}} & \multicolumn{1}{c|}{\scriptsize{Col. 5}} & \multicolumn{1}{c|}{\scriptsize{Col. 6}} & \multicolumn{1}{c|}{\scriptsize{Col. 7}} & \multicolumn{1}{c|}{\scriptsize{Col. 9}}  \\ \hline
    
    LDA         & \multicolumn{1}{c|}{0.8904}  & \multicolumn{1}{c|}{\hspace{0.31cm} 0.9425\hspace{0.31cm} }  & \multicolumn{1}{c|}{\hspace{0.31cm} 0.9297 \hspace{0.31cm} } & \multicolumn{1}{c|}{ \hspace{0.31cm} 0.9240 \hspace{0.31cm}  } & \multicolumn{1}{c|}{ 0.8024 } & \multicolumn{1}{c|}{\hspace{0.31cm} 0.9448\hspace{0.31cm} } & \multicolumn{1}{c|}{0.8503 } & \multicolumn{1}{c|}{\hspace{0.32cm}  0.9591 \hspace{0.32cm}  }  \\ 
    \hline
    
    LR          & \multicolumn{1}{c|}{ \hspace{0.3cm} 0.9497 \hspace{0.3cm} }   & \multicolumn{1}{c|}{ 0.9682}  & \multicolumn{1}{c|}{ 0.8639} & \multicolumn{1}{c|}{ 0.9348 }   & \multicolumn{1}{c|}{ \hspace{0.3cm} 0.8862 \hspace{0.3cm} }    & \multicolumn{1}{c|}{ 0.9387}  & \multicolumn{1}{c|}{ \hspace{0.3cm} 0.9133 \hspace{0.3cm} }   & \multicolumn{1}{c|}{ 0.9653 }  \\ 
    \hline
    
    SVC         & \multicolumn{1}{c|}{0.9371}   & \multicolumn{1}{c|}{ 0.9549}  & \multicolumn{1}{c|}{0.6534 } & \multicolumn{1}{c|}{ 0.9117 }   & \multicolumn{1}{c|}{ 0.7176}    & \multicolumn{1}{c|}{ 0.9570}  & \multicolumn{1}{c|}{0.8276 }   & \multicolumn{1}{c|}{ 0.9527 }  \\ 
    \hline
    
    XGB         & \multicolumn{1}{c|}{ 0.7899}   & \multicolumn{1}{c|}{ 0.9047}  & \multicolumn{1}{c|}{0.9068} & \multicolumn{1}{c|}{ 0.9455 }   & \multicolumn{1}{c|}{0.8917 }    & \multicolumn{1}{c|}{ 0.9598}  & \multicolumn{1}{c|}{0.8774 }   & \multicolumn{1}{c|}{0.9435 }  \\ 
    \hline
    
    ETC  & \multicolumn{1}{c|}{0.5240 }   & \multicolumn{1}{c|}{ 0.8686}  & \multicolumn{1}{c|}{0.9408} & \multicolumn{1}{c|}{ 0.8992 }   & \multicolumn{1}{c|}{ 0.8553}    & \multicolumn{1}{c|}{ 0.8895}  & \multicolumn{1}{c|}{ 0.8701}   & \multicolumn{1}{c|}{ 0.9352 }  \\ 
    \hline
    
    \end{tabular}%}
    \label{tab:modelAdapt}
\end{table*}

\subsection{Feature Aanalysis}
Table~\ref{tab:cheatAUC} reports the F1 score for the different  combinations of features (Section~\ref{sec:features}), obtained by  5-fold cross-validation (CV)  over the \textit{CHEAT} dataset \cite{yu2023cheat}. 

We extracted 115 features in all. Cross-validated performance is reported in Col. 3 of Table~\ref{tab:cheatAUC}, which is fairly high. Combining the three feature sets results into a rich representation of the abstracts and delivers a fairly accurate performance.   %We  conducted an  ablation study by then removing pragmatic features (Hedge words, booster word and hyping lemma) and repeated the experiment. 
The scores in Col. 2 of the table show that the performance degrades by 5 - 10 \% if only linguistic features (LF) are used for training. The extent of loss of performance varies with the classifier, as expected.  This shows the importance of the three pragmatic features.  Finally, we perform cross-validation using only the semantic features (Col. 1), which shows a further loss in performance.  Thus, though both the SF and LF independently contribute significantly to the identification of CG-generated abstracts, the inclusion of pragmatic features improves the performance.
\begin{figure}[ht!]
        \centering
            \begin{subfigure}[b]{0.45\linewidth}
                \centering
                \includegraphics[width=1.5in, height = 1.4in]{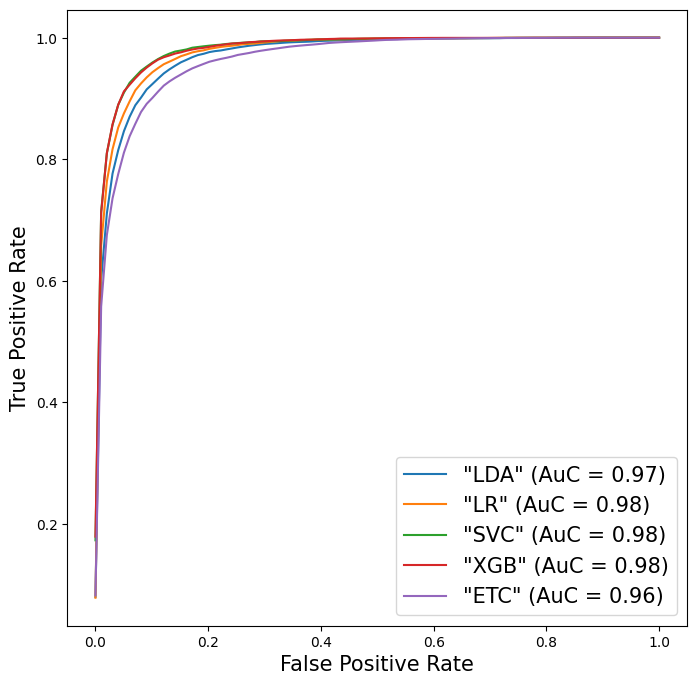}
                \caption{SF + LF + HBH Features}
                \label{fig:cheatAUC_116}
            \end{subfigure}%
            \begin{subfigure}[b]{0.45\linewidth}
                \centering
                \includegraphics[width=1.5in, height = 1.4in]{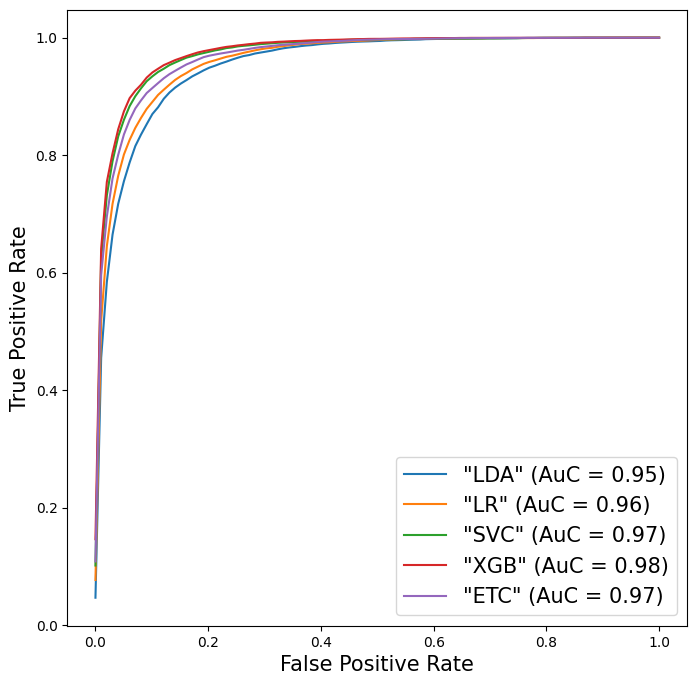}
                \caption{Top-25 Selected Features}
                \label{fig:cheatAUC_25}
            \end{subfigure} 
       \caption{ 5-fold cross-validation performance (AUC) of Machine Learning Models trained on CHEAT dataset.}
        \label{fig:cheatAUC}
\end{figure}
%%%%%%%%%%%% CHEAT FOR TRAINING, HUMAN-EVAL & CG-25 Citations FOR TESTING %%%%%%%%%%%%

In the second experiment, we tested the cross-dataset adaptability of the models trained over \textit{CHEAT} dataset. Table~\ref{tab:modelAdapt} reports the result where the model is tested over \textit{Human-Eval} and \textit{CG-Articles} datasets discussed in Section~\ref{sec:datasets}. Though the test performance on both datasets is fair, we observe that the test performance of the model is slightly better for \textit{CG-Article} dataset compared to \textit{Human-Eval} dataset. This can be attributed to the time of generation of the abstracts. \textit{Human-Eval} dataset was created  in Jan 2023, while \textit{CG-Article} dataset was created in July 2023. Updates in the ChatGPT version may be the reason underlying the difference in the abstracts generated six months apart.
%We expect a small deviation in the feature distribution of human-written scientific abstracts across the years whereas the ChatGPT model is continuously evolving. This may lead to variability in the distribution of several extracted features for the CG-generated abstract. %
\subsection{Feature Selection}
In order to reduce the training cost, we performed feature selection using  XGBoost.   Table \ref{tab:cheatAUC} shows the CV performance of the models on CHEAT dataset. Table \ref{tab:modelAdapt} shows the test performance of the models on top-k selected features. In both cases,  performance improves with the increase in the number of features. We find that CV performance is best on top-25 features. Due to the paucity of space, we omit the list of selected features.

 Area Under the Curve (AUC) for the current experimental setting is reported in Figure~\ref{fig:cheatAUC}. The AUC scores of these simple models are comparable with the result reported by Yu et al.~\cite{yu2023cheat}, where the authors have compared several deep learning-based models for the identification of CG-generated text.  We conclude that classical machine learning algorithms perform reasonably well compared to the predictive performance reported in Table 2 of \cite{yu2023cheat}. Our comprehensive analysis aligns with the recent advocacy by Lin et al.~\cite{lin2023linear} where the authors have shown that simple linear classifiers exhibit comparable performance to that of large-scale pre-trained language models for the text classification task.

%
   
%

%\subsection{Effectiveness of Feature}
     
%\subsection{ Features Ensembling}
    
%\subsection{Measuring Semantic Gap}
 % We measure the semantic gap between CGA and AA using Bert similarity and Earth-mover distance.  We first find the Bert similarity between CGA and AA for all 30 papers and plot the distribution in Fig. \ref{fig:bert-sim}. It is observed that ....ADD... For each paper, we select the CGA with maximum semantic similarity with the author abstract and compute the Earth-mover distance. The distances are plotted in Fig. \ref{fig:em-dist}. It is seen that.... ADD.... 
\section{Conclusion and Future Directions}

We present an empirical study of the language features of human-written and ChatGPT-generated abstracts of scientific scholarly texts. We use linguistic, semantic, and pragmatic features of the text and are able to discriminate the two types of text using \textit{classical machine learning} algorithms with reasonable accuracy.  It is noteworthy that all machine learning models were trained using default parameters. It can be construed that the reported performance can be further boosted by fine-tuning the hyper-parameters of the algorithms. It can also be concluded that a non-linear classifier like Extra Tree achieves a similar accuracy level with approximately 25\% fewer features selected as per the ranking produced by XGBoost. 

One unfinished task in this project is to experiment with features like emotional tones, clout, and analytical thinking for discrimination. These features can be efficiently extracted using LIWC\footnote{\url{https://www.liwc.app/}} software. Another task that remains to be done is to examine qualitative differences between human-written and CG-generated abstracts.

\clearpage
\bibliographystyle{ACM-Reference-Format}
\bibliography{sample-base}
\end{document}